\documentclass[letterpaper, 10 pt, conference]{ieeeconf}  %

\IEEEoverridecommandlockouts                              %

\usepackage{amsmath}
\usepackage{amssymb}  %

\usepackage{bm}  %
\usepackage{graphicx}  %
\usepackage{subcaption} %
\usepackage{siunitx}
\usepackage{flushend} %
\usepackage{wrapfig} %
\usepackage[font=small]{caption} %

\usepackage[
        backend=biber,
        style=numeric,
        sorting=none,
        doi=false,
        isbn=false,
        url=false]{biblatex}

\usepackage{listings}

\title{\LARGE \bf
        Contact Skill Imitation Learning for Robot-Independent Assembly Programming}

\author{Stefan Scherzinger$^{1}$, Arne Roennau$^{1}$ and R\"udiger Dillmann$^{2}$%
\thanks{$^{1}$Stefan Scherzinger and Arne Roennau are with FZI Research Center for Information Technology, Haid-und-Neu-Str. 10-14, 76131 Karlsruhe, Germany
{\tt\small stefan.scherzinger@fzi.de} {\tt\small arne.roennau@fzi.de}}%
\thanks{$^{2}$R\"udiger Dillmann is with IAR Institute for Anthropomatics and Robotics, HIS Humanoids and Intelligence Systems Lab, KIT Karlsruhe Institute of Technology,
        Adenauerring 2, 76131 Karlsruhe, Germany
{\tt\small ruediger.dillmann@kit.edu}}%
}

\begin{document}

\onecolumn
\noindent
\textsuperscript{\textcopyright}2019 IEEE.  Personal use of this material is permitted.  Permission from IEEE must be obtained for all other uses, in any current or future media, including reprinting/republishing this material for advertising or promotional purposes, creating new collective works, for resale or redistribution to servers or lists, or reuse of any copyrighted component of this work in other works.

\vspace{0.5cm}
\noindent
Please cite this paper as:
 
\lstset{%
  basicstyle=\fontsize{9}{10}\selectfont\ttfamily
}%
\begin{lstlisting}
@INPROCEEDINGS{Scherzinger2019Contact,
author={S. {Scherzinger} and A. {Roennau} and R. {Dillmann}},
booktitle={2019 IEEE/RSJ International Conference on Intelligent Robots and Systems (IROS)},
title={Contact Skill Imitation Learning for Robot-Independent Assembly Programming},
year={2019},
volume={},
number={},
pages={4309-4316},
keywords={},
doi={10.1109/IROS40897.2019.8967523},
ISSN={2153-0858},
month={Nov},}
\end{lstlisting}

\twocolumn
\newpage

\maketitle
\thispagestyle{empty}
\pagestyle{empty}

\begin{abstract}
Robotic automation is a key driver for the advancement of technology.
The skills of human workers, however, are difficult to
program and seem currently unmatched by technical systems.
In this work we present a data-driven approach to extract and learn robot-independent
contact skills from human demonstrations in simulation environments,
using a Long Short Term Memory (LSTM) network.
Our model learns to generate error-correcting sequences of forces and torques in
task space from object-relative motion, which industrial robots carry out through
a Cartesian force control scheme on the real setup.
This scheme uses forward dynamics computation of a virtually conditioned twin of
the manipulator to solve the inverse kinematics problem.
We evaluate our methods with an assembly experiment,
in which our algorithm handles part tilting and jamming in order to succeed.
The results show that the skill is robust towards
localization uncertainty in task space and across different joint configurations of the robot.
With our approach, non-experts can easily program force-sensitive assembly tasks in a robot-independent way.
\end{abstract}

\section{INTRODUCTION}  %

The robotic automation of assembly tasks is one of the oldest fields of robotic
applications and has challenged engineers and scientists ever since.
In the domain of robotic assembly, having generalized strategies
has been a strong incentive, since it would enable to program
many robots in short time by effectively re-using solutions.
However, highly varying system dynamics
usually require specific strategies to deal with part tilting and jamming during
execution, and are difficult to generalize to other robots and work postures.
In this work,
we propose a data-driven approach to obtain contact skills that
encapsulate human manipulation strategies that we extract from demonstrations in simulation.
Robots can then execute these skills with a common force-control interface.

Flexible automation and intelligent robots in industry have been considered for decades to
be substantial for industrialized countries \cite{Nitzan1976}, \cite{Nitzan1985}.
Towards this goal, works investigated the mechanics of
assembly to derive general analytic solutions and principles, e.g. to handle friction \cite{Schimmels1993}, part jamming that arise through sensor imprecisions \cite{Whitney1982}, or targeted planar parts with 
compliance parameter optimization \cite{Stemmer2007}.
Contacts and contact transitions have been investigated between manipulation objects \cite{Xiao2001}, \cite{Tang2006}, whose semantic information can speed-up the design of compliant motion
for robot end-effectors in contact with their environments \cite{Meeussen2007contact}.
Following the \textit{task-level} approach, works used
primitives to compose \textit{skills}, e.g. by formulating position-force
commands, such as 'rotate to level', 'rotate to insert' \cite{Hasegawa1992}, or
concatenating sensorimotor primitives \cite{Morrow1995}, \cite{Morrow1997},
using human inspired recipes \cite{Newman1999} or formulating elementary
actions in the task frame \cite{Bruyninckx1996}.  The idea of elementary skills
has still gained some interest in recent works \cite{Wahrburg2015},
\cite{Halt2018}.
However, the ingenuity of finding the primitives that apply best in certain
situations and parameterizing them often means a considerable engineering
effort, leaving the cognitive performance to the programmer and not to the system.
Humans are remarkably skilled workers and join assembly parts with ease, although not necessarily using
analytical representations, such as contact states \cite{Klingbeil2016}.
However, it is not intuitive for us to describe how we deal with tilting and
jamming or what strategies we deploy upon getting stuck.
Following this insight, human performance can directly be used to obtain skills through imitation learning, such as in \textit{Programming by Demonstration} (PbD)
\footnote{Also Learning from Demonstration (LfD)}
\cite{Dillmann1995}, \cite{Tung1995}, \cite{Dillmann2004}, 
with applications for general object manipulation \cite{Jaekel2010}, \cite{Kormushev2011}, \cite{Rozo2013},
and industrial assembly processes \cite{Krueger2014}, \cite{Savarimuthu2017}.
Contrary, approaches have shown promising results on contact-rich manipulation tasks without human input
\cite{Levine2015Learning},\cite{Levine2016End}, also with very tight clearances \cite{Inoue2017}.
Transferring the solutions to arbitrary poses in the workspace or even to other
systems, however, requires the
training of a new controller, even if initial trajectories are provided
\cite{Thomas2018}.

In this work, we aim at developing force-based contact skills to handle jamming
and tilting effects that are portable to different robotic manipulators,
and can, once learned, serve as robot independent skills for a specific
assembly task.
To this end, we train a recurrent neural network to learn human-like
manipulation strategies from human performance in simulation, and relate those
to the relative object's geometry in task space.
In contrast to related works, our model predicts sequences of forces and
torques, which serve as reference set point for a Cartesian force-control
of position-controlled robots, which we build upon the idea from an earlier work \cite{Scherzinger2017}.

The remainder of the paper is as follows:
In \ref{sec:related_work} we discuss related work and motivate our approach,
which we explain in detail in \ref{sec:model} for our contact skill model,
and in \ref{sec:robot_control} for the robot-abstracting force-control.
\ref{sec:results} shows our experiments and results. In~\ref{sec:discussion} we discuss final aspects
and conclude in~\ref{sec:conclusions}.
\section{RELATED WORK}
\label{sec:related_work}

Imitation learning of human assembly skills has been tackled with various
approaches, such as using teleoperation \cite{Yang1994}, \cite{Krueger2014},
\cite{Savarimuthu2017}, teaching with a haptic device \cite{Skubic2000},
direct kinesthetic manipulator guiding \cite{Kramberger2016} or the explicit usage of
simulation to record human performance, \cite{Ogata1994}, \cite{Onda1997},
\cite{Onda1995}, \cite{Dong2007}.
Our work incorporates both the idea of teleoperation for human skill extraction
and the advantage of simulation to acquire numerous samples in short time,
using Long Short-Term Memory (LSTM) \cite{Hochreiter1997} as a method to
model assembly sequences. 
A similar approach has been used by Rahmatizadeh et al~\cite{Rahmatizadeh2018} to learn basic
manipulation skills with a gamepad, albeit not
explicitly targeting contact skill dominated scenarios or robot transfer.

Early proposals for the usage of simulation are from Ogata and Takahashi
\cite{Ogata1994} and Onda et al \cite{Onda1995}, \cite{Onda1997}, in which they
used a robotic manipulator as teach device to steer objects in simulation with
the aim to extract the visiting of contact states during assembly. They later
derived hybrid position/force commands, using the work of Hasegawa et al
\cite{Hasegawa1992}.  Skubic et al~\cite{Skubic2000} learned mappings from sensor forces to
contact formations directly, obtaining samples via a haptic device.
After classifying sequences of those formations, they
commanded the robots with velocities using a finite state machine.  In a similar
approach, Dong et al \cite{Dong2007} relied on performance in a virtual
environment to identify the contact states with a Hidden Markov Model (HMM).
They used Locally Weighted Regression (LWR) to learn rotation angle correction
trajectories for a 3 DOF (Degree of Freedom) task.

Krueger et al \cite{Krueger2014}, and later Savarimuthu et
al \cite{Savarimuthu2017} used magnetic trackers in the assembly objects to
record Cartesian trajectories during assembling the Cranfield benchmark set.
The robot imitated the human performance in teleoperation mode on a twin of the assembly setup, while
recording the forces with the robot's end-effector sensor.
Instead of requiring special setup for the skill extraction, Kramberger et al~\cite{Kramberger2016}
directly guided a light-weight robot during the task,
recording Cartesian trajectories and force profiles, which were then
generalized, using Locally Weighted Regression (LWR).

In contrast to those works, we steer the assembly objects directly with
forces and torques, both during the acquisition of training data and during the
execution on the robotic system. There are two main reasons for this choice:
First, learning mappings from object's geometry to sequences in wrench space
allows us to cope with various disturbances (self-provoked or external), by
continuing our work from multiple entrance points.
Ideally, we plan to recover from arbitrary, relative object poses.
Second: Commanding in wrench space allows us to scale the output of our
model easily without changing its semantics.
Both require an interface to gravity-compensated force control on the robot as described in section \ref{sec:robot_control}.

\section{MODEL}       %
\label{sec:model}

\subsection{Skill extraction}
\label{sec:skill_extraction}
\begin{figure}
        \centering
        \begin{subfigure}[b]{0.49\textwidth}
                \label{fig:gazebo}
                \centering
                \includegraphics[width=1.0\textwidth]{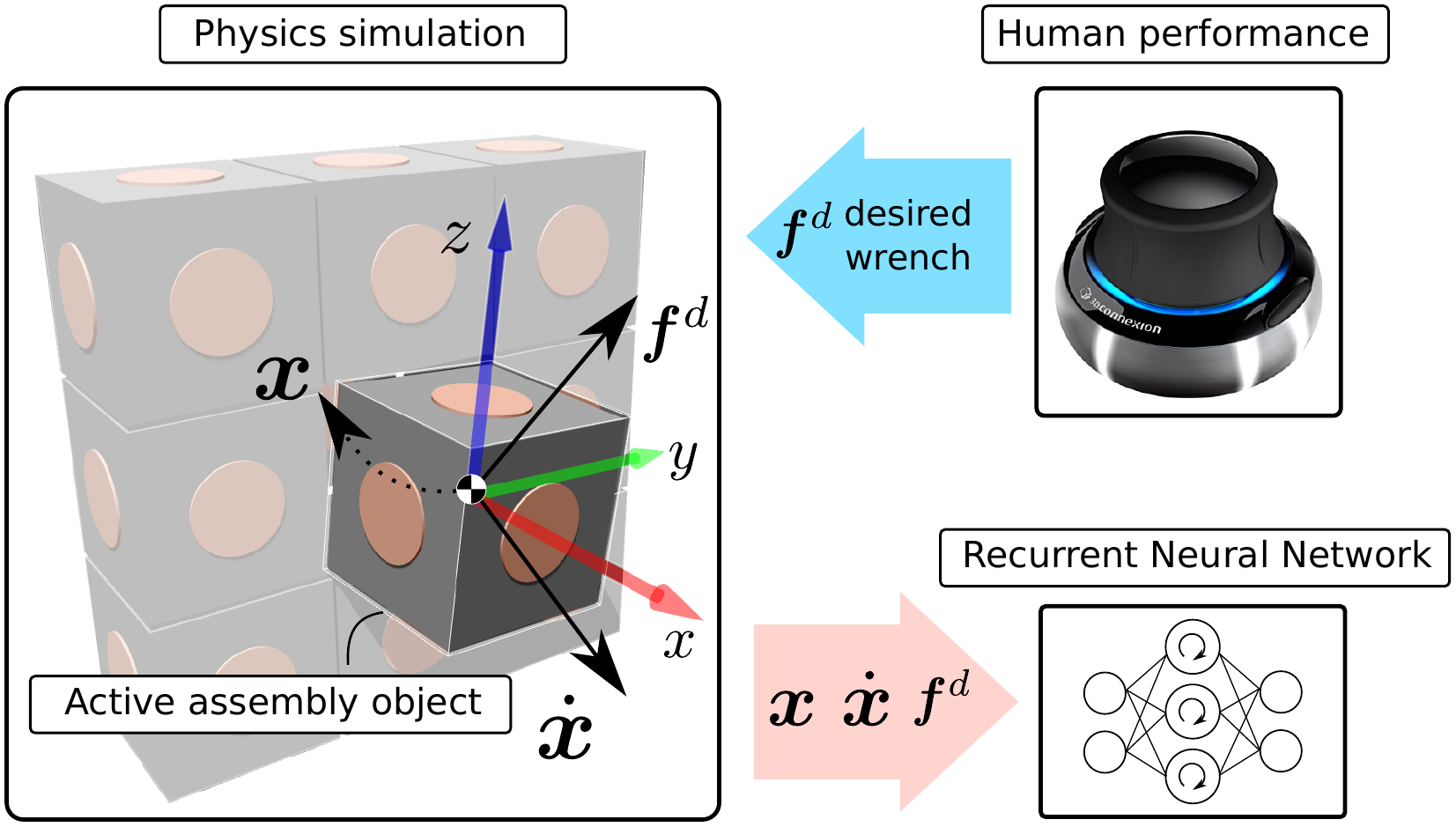}
                \caption{}
        \end{subfigure}
        \begin{subfigure}[b]{0.45\textwidth}
                \label{fig:tilting_and_jamming}
                \centering
                \includegraphics[width=1.0\textwidth]{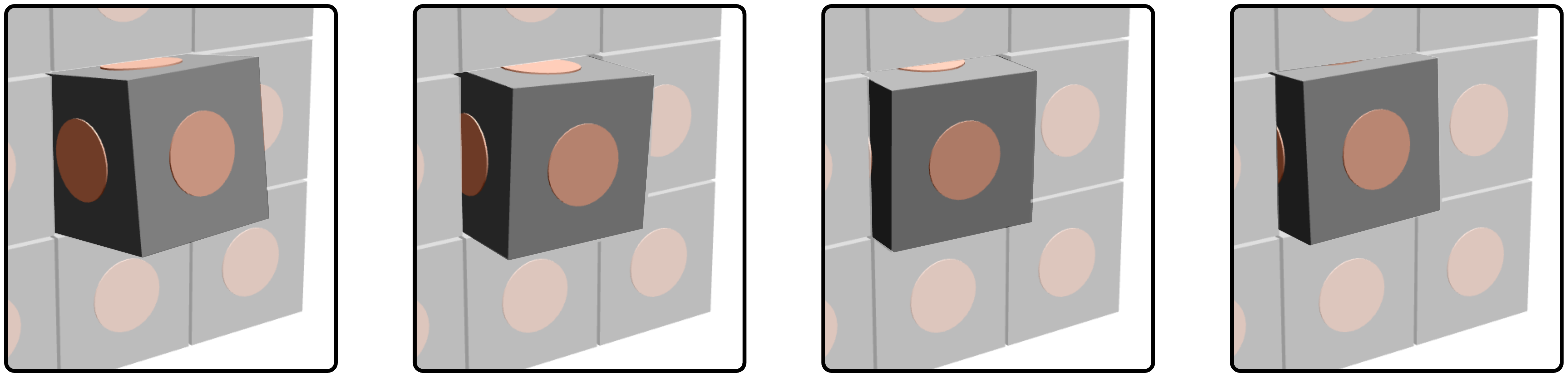}
                \caption{}
        \end{subfigure}
        \caption{Skill extraction. (a) Users solve the task by steering the objects with a teach device, using the simulation's rendered world as feedback for their actions. (b) For a successful assembly, users must correct tilting and jamming of the parts that occur due to friction and small clearances of the setup.}
        \label{fig:training_data_generation}
\end{figure}

We consider the task of learning assembly skills from human performance in simulation.
To this end, we assume that the geometry of all objects involved in the task is known, and we model
the assembly parts as rigid bodies with collision physics in a zero gravity environment
as shown in Fig.~\ref{fig:training_data_generation} for an exemplary box assembly.
We find that the features mentioned are common place for most robot simulation environments, such as Gazebo \cite{Koenig2004}, which we use for our simulations.
Note that our approach does not necessarily require realistic mass nor inertia
parameters of the objects, which, in contrast to geometry, could be
difficult to obtain.  Approximate values are sufficient, as
long as steering the objects in simulation feels natural and responsive enough.
For the active object, we impose a velocity
proportional damping according to
\begin{equation}
        \label{eq:simulation_damping}
                \begin{pmatrix}
                        \bm{I}^{3 \times 3} d_{\text{lin}} & \bm{0} \\
                        \bm{0} & \bm{I}^{3 \times 3} d_{\text{rot}}
                \end{pmatrix}
        \dot{\bm{x}} + \bm{f}^d  + \bm{f}^c = \bm{0}
\end{equation}
in which $\dot{\bm{x}} = \left[ \dot{x}, \dot{y}, \dot{z}, \dot{r}_x,
\dot{r}_y, \dot{r}_z \right]^{T}$ is the floating part's 6-dimensional velocity with linear and
rotary components, $d_{\text{lin}}$, $d_{\text{rot}}$ are linear and rotary damping,
$\bm{f}^d = \left[ f_x, f_y, f_z, t_x, t_y, t_z \right]^{T}$
is the wrench of forces and torques applied to its center of mass by the user via the teach device and
$\bm{f}^c$ is a physics engine controlled wrench to maintain contact stability
and prevent objects from penetrating.

While applying forces and torques to objects is an essential feat of
physics engines, it is commonly difficult to obtain realistic
contact forces of those objects with virtual sensors, due to approximations of
friction for rigid bodies and numerical constraint instabilities.
Our idea was therefore to avoid the
dependency of those forces as input features to our neural network:
The users correct the outcome of
their force-based actions through observing the rendered world in simulation.
We did not include any other visual feedback, such as plots of force-torque
readings.
We assume that the users' demonstrations in this visual servoing approach
contain sufficient semantic information to learn error-correcting skills relative to the objects.
Through this approach, we aim to prevent depending on force-feedback as input feature that
would be prone to suffering the "reality gap" when working on the real system.
Still relying on simulation provides us with a method to carry out a big number of
trials as easy as possible to generate sufficient data for our deep network.

To this end, users steer the active assembly parts with the help of a teach
device, in our case a conventional space mouse, which we deploy as a sensor for 6-dimensional motion. We map
those inputs to forces and torques, such that offsets to the sensor's initial position scale linearly with the
magnitude of the wrench $\bm{f}^d$ applied to the object in simulation.
We continuously record $\bm{f}^d$ along with the target pose $\bm{x} = \left[ x,y,z,q_x,q_y,q_z,q_w \right]$ of the
final assembly operation with the orientation given in quaternion notation, and the objects 6-dimensional velocity $\dot{\bm{x}}$.
Note that we transform and display all quantities with respect to the
moving frame of the active assembly object.
The user initiated commands $\bm{f}^d$ represent expert behavior in each situation, which
entail both micro strategies with short time horizon against getting stuck at edges and
macro strategies with longer time horizon for more path planning behavior.
\subsection{LSTM-based Contact Skill Models}
\begin{wrapfigure}{r}{0.15\textwidth}
        \includegraphics[width=0.15\textwidth]{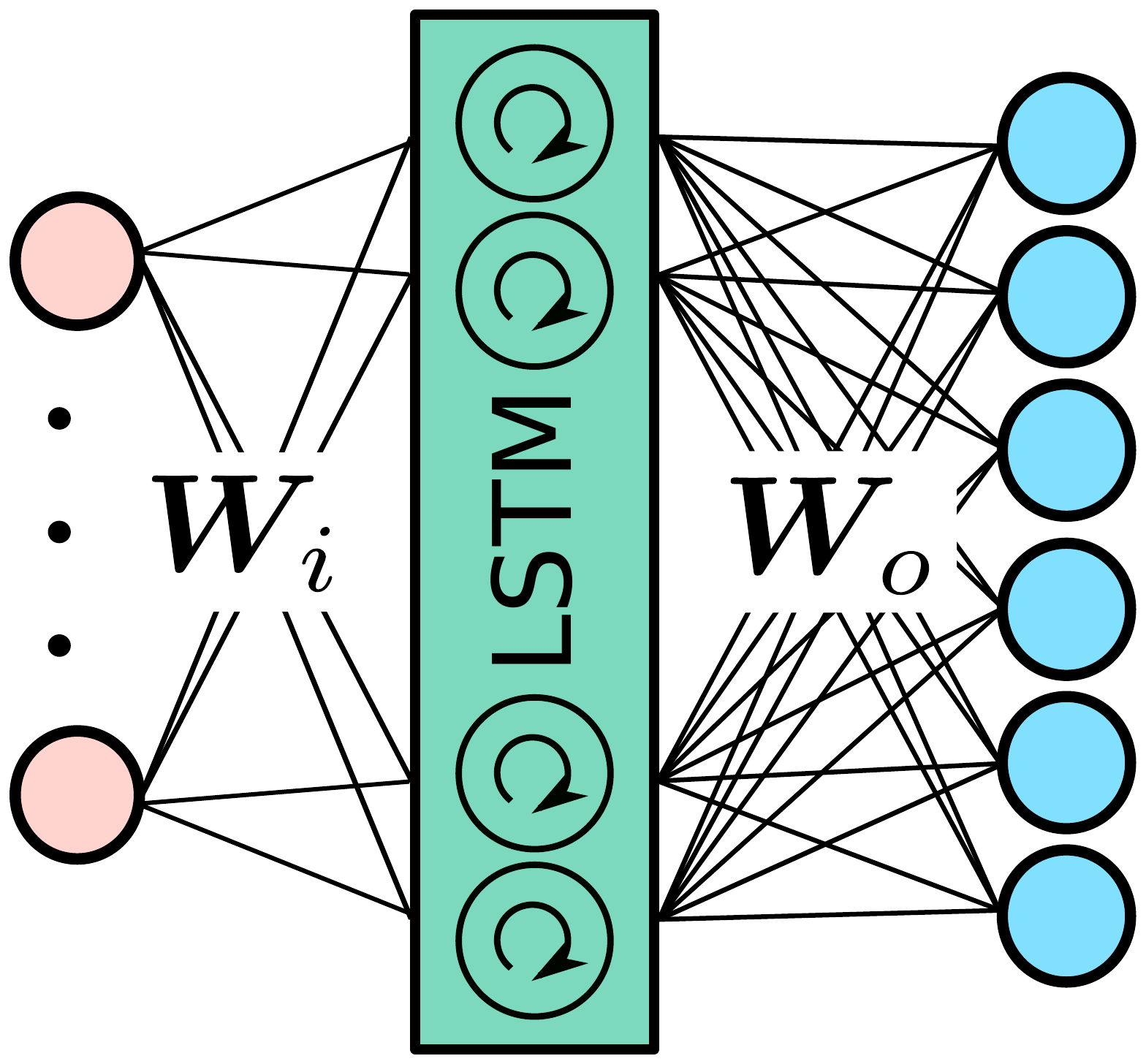}
        \caption{Skill model}
        \label{fig:skill_thumbnail}
\end{wrapfigure}

\begin{figure}
        \centering
        \includegraphics[width=.45\textwidth]{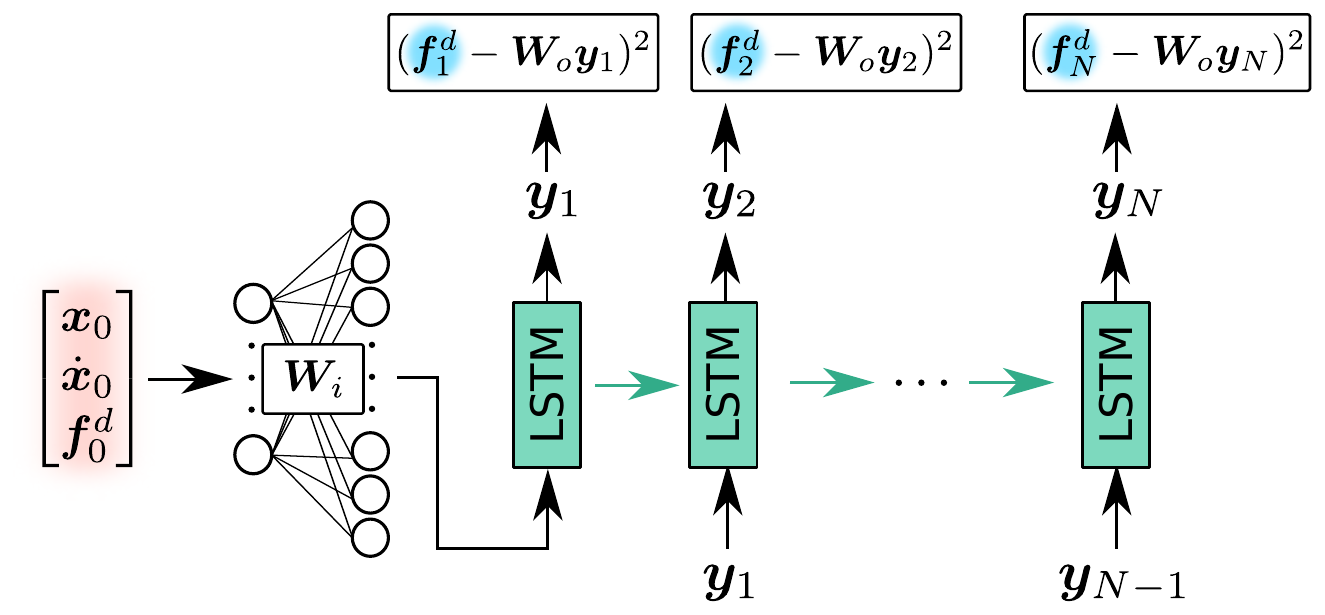}
        \caption{Unrolling the neural network over time.}
        \label{fig:model}
\end{figure}

We use an LSTM-based model (Fig.~\ref{fig:skill_thumbnail}) to learn and generalize human skills in form of
\textit{one-to-many} mappings.
In contrast to feed forward networks, LSTM cells keep an internal state,
enabling them to learn across various time steps.
More details can be found in the original work \cite{Hochreiter1997} and the
refinement \cite{Gers2000}, which is also the base for the implementation we
use.
When using them recursively on their own predictions,
LSTMs have been shown to produce creative sequences from single inputs \cite{Vinyals2015}.
Although residing in another domain, our application has similarities to this approach:
In our work, to a given seed input - an estimated state in the middle of an assembly operation - our
model should generate a creative sequence that is representative of human behavior in that
scenario.
Our idea is to drop the model into a specific situation and let it predict a meaningful sequence of next steps in form of forces and torques $\bm{f}^d$.
The seed input features are composed as the tuple $\left[ \bm{x}_0 \ \dot{\bm{x}}_0 \ \bm{f}^d_0 \right]$.
where the subscript ${}_0$ shall reflect any starting point in time from which on a sequence is to be predicted by our model.
We unroll our network over a number of fixed steps $N$ both during training and inference.
The unrolling procedure is as follows:
\begin{equation}
        \label{eq:lstm_unrolling}
        \begin{aligned} 
                \bm{y}_{0} & = 
                \bm{W}_i 
                \left[
                        \bm{x}_{0},
                        \dot{\bm{x}}_{0},
                        \bm{f}^d_{0}
                        \right]^T \\
                \bm{y}_{k} & = \text{LSTM}(\bm{y}_{k-1}) , \qquad k \in \{1,..,N\} \\
                \bm{p}_{k} & = \bm{W}_o \bm{y}_k , \qquad k \in \{1,..,N\}
        \end{aligned}
\end{equation}
Note that we use the model's prediction in each step as the next input, as is shown in Fig.~\ref{fig:model}.
In all unrolled steps, the LSTMs share the same parameters, which we optimize together with the input weights $\bm{W}_i$ and output weights $\bm{W}_o$ subject to our loss
\begin{equation}
        \label{eq:mse}
        \bm{L}(\bm{x}_0, \dot{\bm{x}}_0, \bm{f}^d_0, \bm{f}^d) =
                \frac{1}{N} \sum^{N}_{k=1} ( \bm{f}^d_k - \bm{p}_k )^{2} .
\end{equation}

\subsection{Training}
\begin{figure}
        \centering
        \includegraphics[width=.5\textwidth]{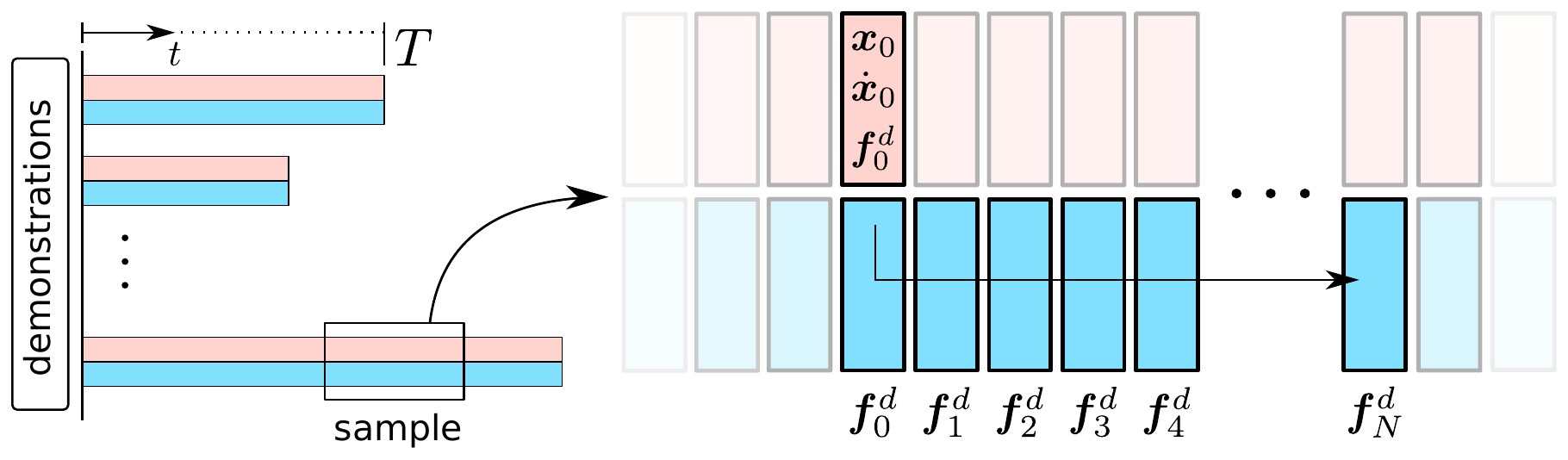}
        \caption{Composition of training samples.}
        \label{fig:composition_training_samples}
\end{figure}

Fig.~\ref{fig:composition_training_samples} shows the composition of training
samples.
The records from simulation and readings from the teach device form a multitude
of demonstrations, each with an individual temporal length $T$.  The difference
in length is due to non-deterministic human performance throughout the task and
 differences in the random starting poses of the active assembly object.
 Although the sequences depicted in light blue decode expert behavior, the commands issued
 by users are not optimal and
likely contradictory to certain degree.  We assume, however, that they
statistically contain sufficient consensus on "the correct behavior" in jamming
situations.  During training, we take samples randomly from the total of
demonstrations.  Each sample is comprised of the input seed and the following
human performance as sequence of labels.  Note that the time span of the sample
length $N$ is smaller than an individual complete demonstration $T$.  We train the
network with Back Propagation Through Time (BPTT), using mini-batch
stochastic gradient descent.

\section{ROBOT CONTROL}
\label{sec:robot_control}
\begin{figure*}[htbp]
        \centering
        \includegraphics[width=0.85\textwidth]{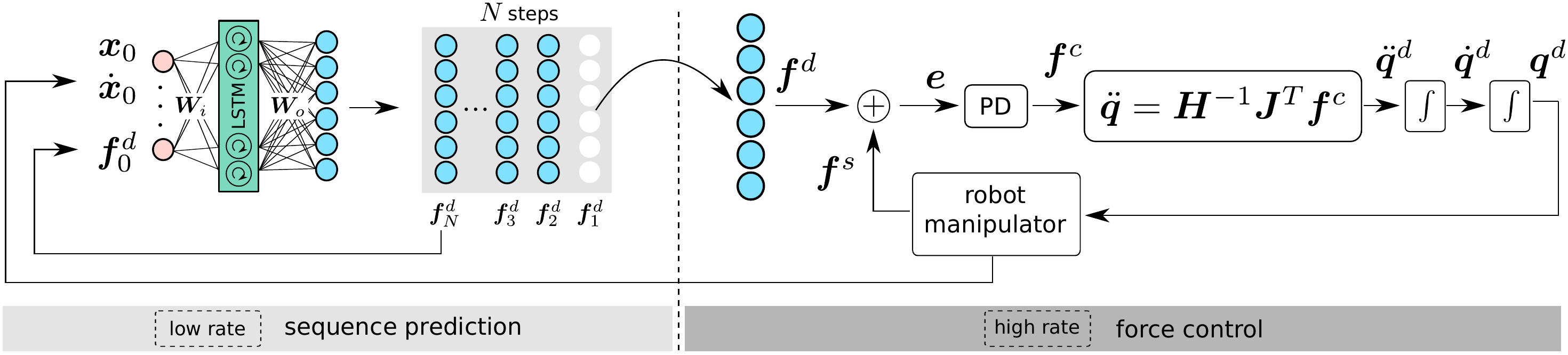}
        \caption{Skill model in combination with Cartesian force control. The neural network
        predicts a sequence of $N$ steps, whose individual elements
        $\bm{f}^d$ serve as reference setpoint for the force regulating PD controller.
        The plant uses realistic manipulator kinematics $\bm{J}$, and a
        virtually conditioned manipulator inertia matrix $\bm{H}$.
        }
        \label{fig:force_control}
\end{figure*}
We consider the class of robots with high-gain position servo in the joints,
which provide an interface for commanding feed forward joint trajectories $\bm{q}^{d}(t)$ with
individual reference set points.
We further assume that our robots possess sensors to measure the wrench $\bm{f}^{s}$ at
their end-effectors, i.e. at the grasped assembly objects and that they provide
functionality to bias the sensors. This functionality is important for us in
order to replicate the zero gravity environment we had chosen during the skill
extraction in simulation.
Relying on additional sensors is not restrictive, given that 6-axis
force-torque sensors have become widely available for industrial robots.
To derive our control concept, we start with the common equations of motion for
an articulated rigid body system in joint coordinates $\bm{q}$
\begin{equation}
        \label{eq:rigid_body_system}
        \bm{H} \left(\bm{q}\right) \ddot{\bm{q}} + \bm{C} \left( \bm{q},\dot{\bm{q}}\right) = \bm{\tau}_{u} + \bm{J}^{T} \bm{f}^{ext}
\end{equation}
where $\bm{H}(\bm{q})$ denotes the positive definite joint inertia matrix,
$\bm{C}$ encompasses centrifugal and Coriolis terms, as well as gravitational forces, and
$\bm{\tau}_{u}$ denote the
motor torques in the joints of the manipulator.
External forces and torques $\bm{f}^{ext}$, acting on the end effector map to joint space with the manipulator Jacobian $\bm{J}(\bm{q})$.
We skip the dependency of $\bm{q}$ in further notation for brevity.
Under the assumption that we move slowly in contacts and that the robots we
consider already compensate for gravity in their lower level control, we omit
$\bm{C}$ and $\bm{\tau}$, such that we obtain
\begin{equation}
        \label{eq:simplified_system}
        \ddot{\bm{q}} = \bm{H}^{-1} \bm{J}^{T} \bm{f}^{ext}
\end{equation}
as an instruction to move according to external disturbances.
In contrast to our previous work \cite{Scherzinger2017} we use the
approximation of (\ref{eq:simplified_system}) for the quasi static case.
This mapping computes the instantaneous joint acceleration of our virtual
system, reacting in direction of our wrench vector $\bm{f}^d$.
While $\bm{J}$ reflects our real system's kinematics, we
chose $\bm{H}$ by setting the masses of all links to some unit values.
Note that we do not estimate the real system dynamics.
Instead, we follow the idea from \cite{Scherzinger2017}, and deploy
(\ref{eq:simplified_system}) as a \textit{forward dynamics solver} for the
inverse kinematics problem on a virtually conditioned twin of the real manipulator.
After double time integration we obtain the new joint commands $\bm{q}^d$, which we
sent open-loop to the black box control of the robot, whose
low-level joint position servo compensates for dynamics-introduced disturbances and gravity.
The force-torque sensor's readings close the control loop in contacts with the environment.
We discuss limitations of this black box restriction briefly in section \ref{sec:discussion}.

Fig.~\ref{fig:force_control} depicts the control scheme, in which
$\bm{f}^d$ is the feed forward prediction of the neural network at each time step and
$\bm{q}^d$ is the joint position set point for the robot.
The \textit{PD} controller regulates $\bm{f}^c$ according to $f^c = k_p e + k_d \dot{e}$
for each of the six Cartesian components until $\bm{f}^d$ is equilibrated by
the individual measured components in $\bm{f}^s$. The gains $k_p$ and $k_d$
are, in combination with the chosen dynamics of $\bm{H}$, a partly redundant means to
adjust the systems responsiveness.
Our overall incentive is to achieve that the assembly objects respond equally to $\bm{f}^d$, both during
supervised training in simulation (without manipulator) and during neural network controlled execution on real robotic manipulators.

\begin{figure}
        \centering
        \includegraphics[width=.4\textwidth]{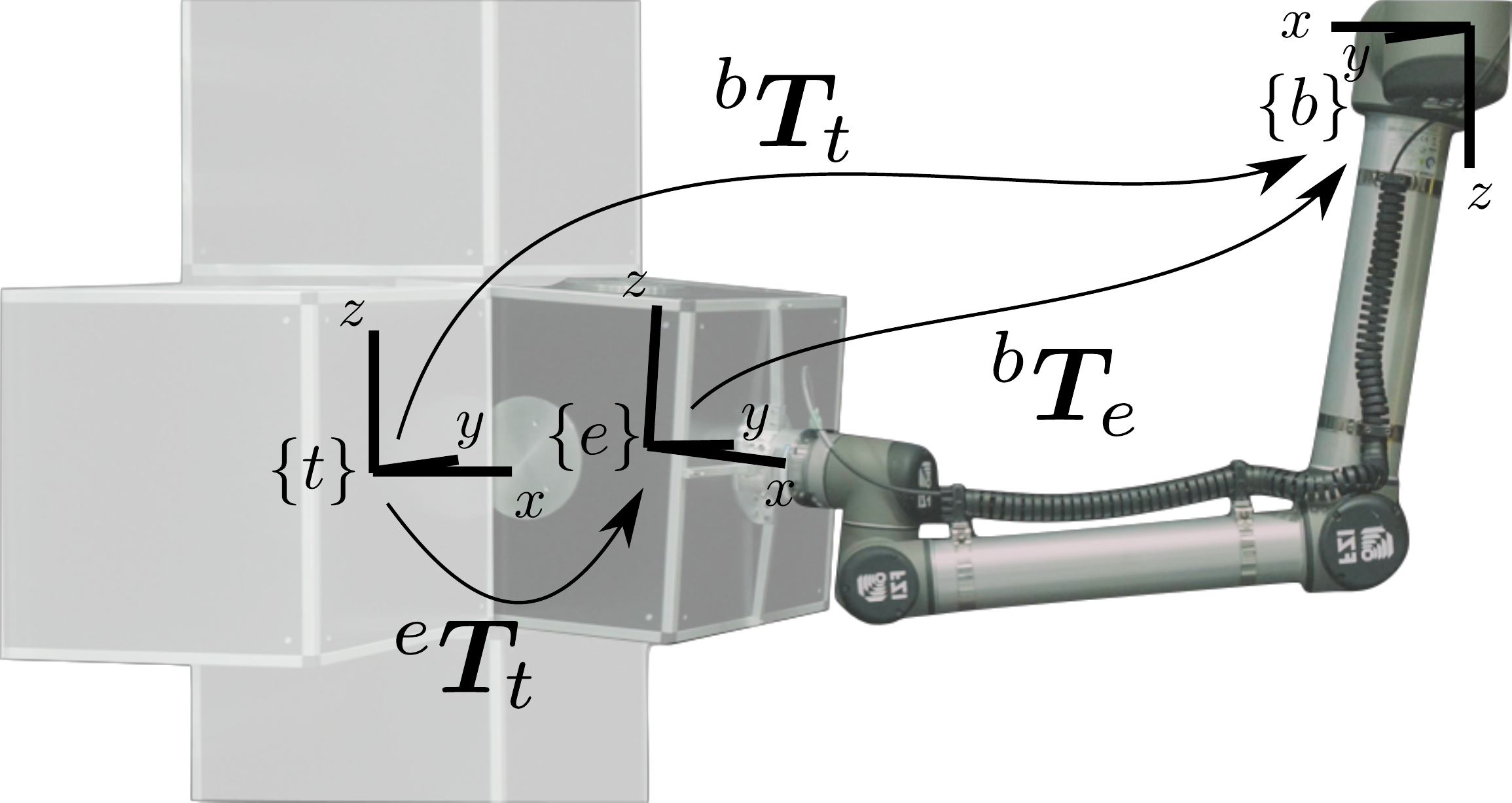}
        \caption{Coordinate systems and transformations to describe task and robot relations.}
        \label{fig:transformations}
\end{figure}

We compute the input features for the neural network using joint state feedback $\bm{q}, \dot{\bm{q}}$ from the robot.
Fig.~\ref{fig:transformations} shows the reference frames and relationships. In the following
notation, the sub- and superscripts $t$, $e$ and $b$ stand for \textit{target},
\textit{end-effector} and \textit{base} respectively.
We assume that an estimation of the final target pose $\{t\}$ of the assembly
operation is given with respect to the robot base $\{b\}$, which we denote with
the homogeneous transformation $^{b}\bm{T}_t$.  Its ground truth will coincide
with the robot's end effector
pose $^{b}\bm{T}_e$ after a successful task execution.
The current end-effector pose is computed according to $^{b}\bm{T}_e =
g(\bm{q})$ at every time step, using the position sensors readings $\bm{q}$ and
a \textit{forward kinematics routine} $g$, which we do not further specify.
Using the inverse of this transform $(^b \bm{T}_e)^{-1} = {}^e \bm{T}_b$, the estimated target pose can be formulated with respect
to the moving end-effector frame according to $^e \bm{T}_t = {}^e\bm{T}_b {}^b \bm{T}_t$, from which it is straight-forward to extract our input features in quaternion notation.
Additionally, we compute the current end-effector velocity $\bm{J}
\dot{\bm{q}}$, and likewise display the solution seen from the end
effector frame, using $^e\bm{R}_b$ as the pure rotational part of $^e\bm{T}_b$.

Deriving the input features for our neural network summarizes as follows:
\begin{equation}
        \label{eq:input_features}
        \begin{aligned} 
                \left[x,y,z,q_x,q_y,q_z,q_w\right]^T &\leftarrow ^e\bm{T}_t \\
                \left[
                        \dot{x},
                        \dot{y},
                        \dot{z},
                        \dot{r}_x,
                        \dot{r}_y,
                        \dot{r}_z
                        \right]^T &\leftarrow 
                \begin{pmatrix}
                        ^e\bm{R}_b & \bm{0} \\
                        \bm{0} & ^e\bm{R}_b
                \end{pmatrix}
                \bm{J} \dot{\bm{q}} \\
                \left[f_x,f_y,f_z,t_x,t_y,t_z\right]^T &\leftarrow \bm{f}^d
        \end{aligned}
\end{equation}

\section{EXPERIMENTAL RESULTS}       %
\label{sec:results}
We conducted a set of experiments to evaluate our approach of contact skill
imitation learning in simulation with transfer to a real robot.

\subsection{Implementation and setup}
We implemented our neural network model in Python, using the machine learning framework
Tensorflow \cite{Tensorflow2015}, and implemented the force control from Fig.~\ref{fig:force_control}
in C++ as a real-time ROS-controller for the ROS-control framework \cite{Chitta2017}.
We designed the test setup to include spots for form-closure effects, such as
collisions and jamming during insertion. We realized this through a set of cube
structures with round plates on the sides, as illustrated in
Fig.~\ref{fig:jamming_skills}.
Table~\ref{tab:setup} summarizes the details of the task setup.
\begin{table}
        \caption{Specifications of the assembly task setup}
\begin{center}
        \begin{tabular}[htbp]{ | c | c | c | c | c |}
                \hline
                cube &
                \multicolumn{2}{c|}{side plates} &
                \multicolumn{2}{c|}{mockup clearance} \\
                \hline
                edge length & thickness & diameter & trans. & rot. \\\relax
                [mm] & [mm] & [mm] & [mm] & [deg] \\
                \hline
                400 & 4 & 145 & 2 & 0.8 \\
                \hline
        \end{tabular}
\end{center}
\label{tab:setup}
\end{table}

We further chose the Universal Robots UR10 as exemplary platform for the task transfer, which is a common
place, joint position-controlled, industrial robot.
It provided both the reach and the end-effector load capacity for our task.
\subsection{Contact skill learning}
\begin{figure}
        \centering
        \includegraphics[width=.48\textwidth]{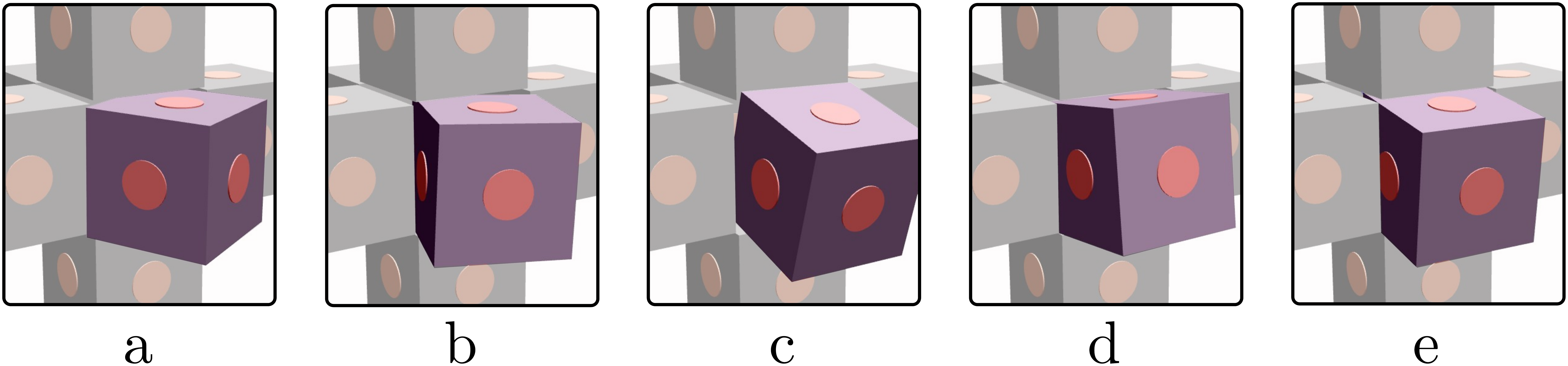}
        \caption{Starting poses for the experiments in simulation. No manipulator is involved in these experiments.}
        \label{fig:jamming_skills}
\end{figure}

\begin{figure}
        \centering
        \includegraphics[width=.5\textwidth]{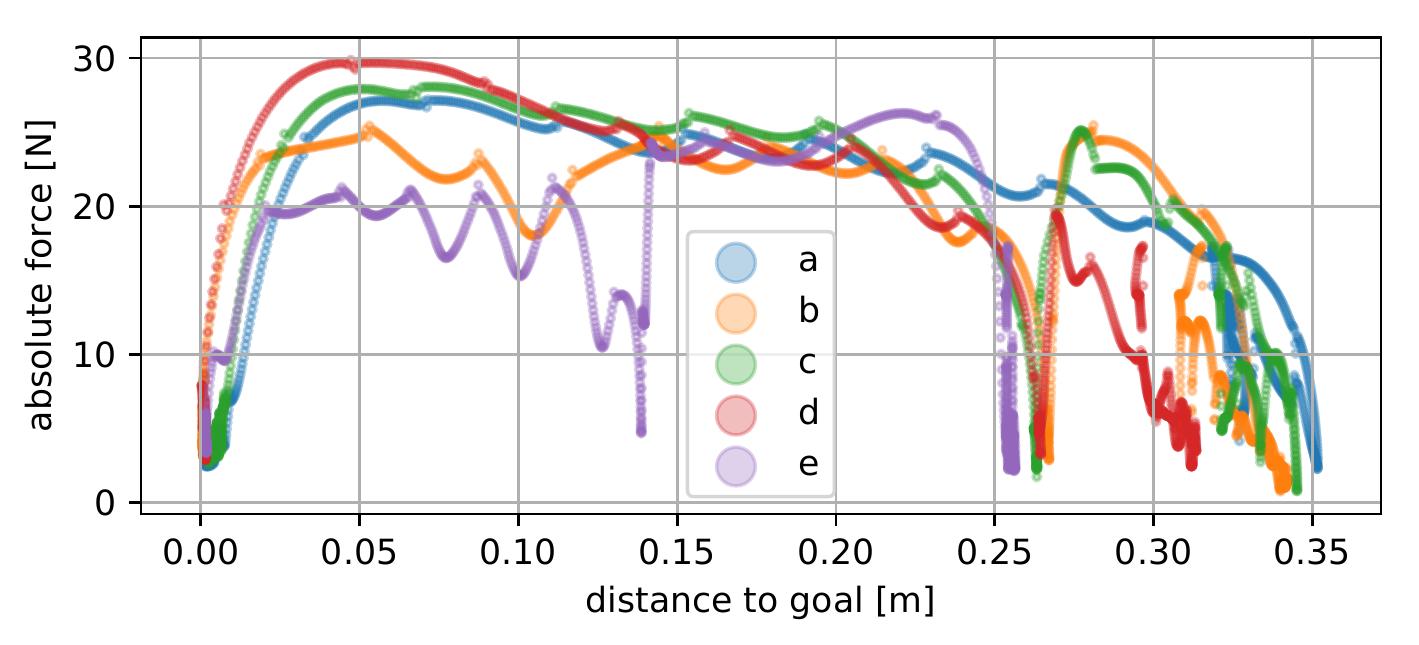}
        \caption{Absolute forces applied by our model along the way to the goal
        pose. The curves a - e correspond to the starting poses of
        Fig.~\ref{fig:jamming_skills}}
        \label{fig:jamming_forces}
\end{figure}

\begin{figure}
        \centering
        \includegraphics[width=.5\textwidth]{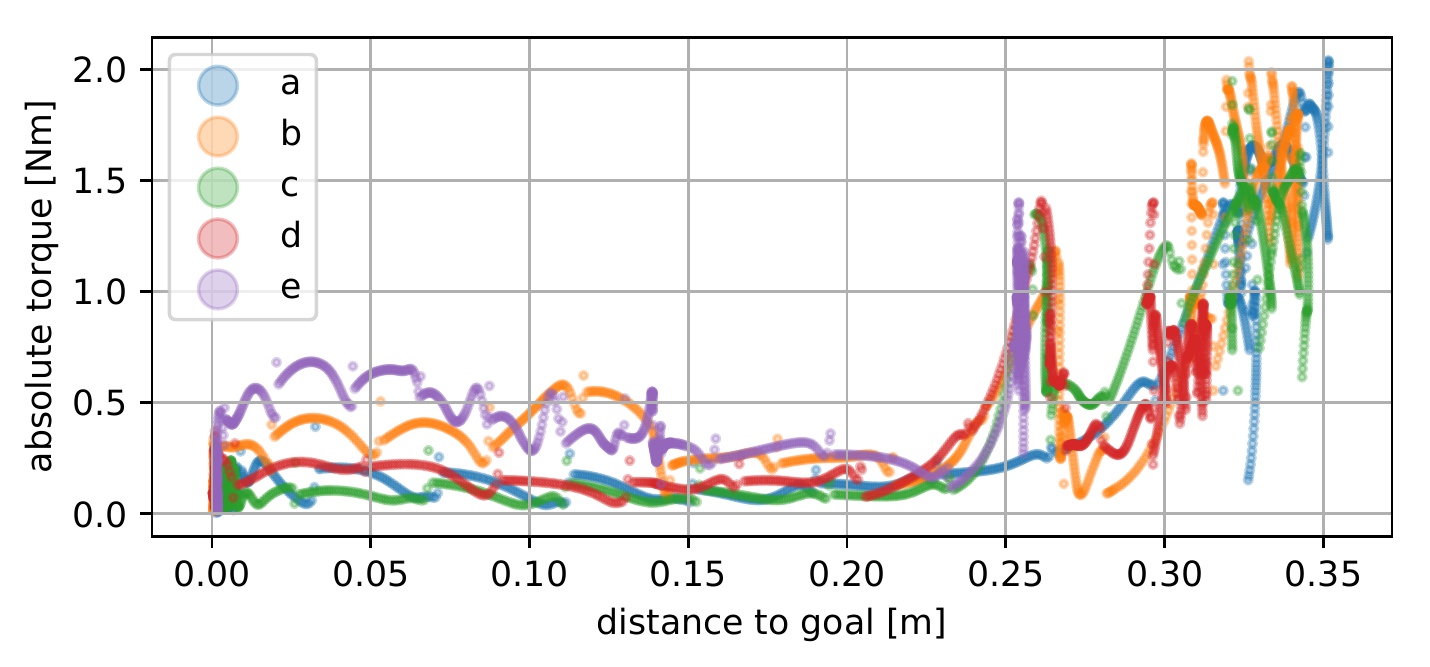}
        \caption{Absolute torques applied by our model. The curves a - e correspond to the starting poses of
        Fig.~\ref{fig:jamming_skills}}
        \label{fig:jamming_torques}
\end{figure}

The training samples for the task of our experiments were generated in the simulation environment as described in \ref{sec:skill_extraction}, using a logging rate of \SI{100}{Hz}.
We trained on approximately 1000 demonstrations, each representing a successful
insertion by a human expert, classically lasting between \SI{10}{s} and \SI{15}{s}, depending on the random starting poses.
For all experiments we used 50 cells in our LSTM layer, a mini-batch size of 512 and sequences of $N = 50$
steps in BPTT, corresponding to time slices of \SI{0.5}{s} in the training set.
We applied Dropout \cite{Srivastava2014} as regularization technique to prevent our network from overfitting, and
used Adam \cite{Kingma2014} as optimizer.
We tested the learned skills with bringing
the neural network in a multitude of unseen jamming situations in simulation and letting it solve the task.
Five of these starting poses are depicted in Fig.~\ref{fig:jamming_skills}.
The purple cube was oriented in such a way that it was mandatory for the model
to retreat and to apply a more sophisticated strategy than simple goal-directed pushes
to succeed.
We set the serving rate of the neural network to \SI{50}{Hz}, covering a time span of \SI{1}{s} in simulation.
Fig.s~\ref{fig:jamming_forces} and \ref{fig:jamming_torques} show the course
of the forces and torques as applied by our model along the Euclidean distance to each goal position (reading from right to left).
We found that the forces rapidly increased at the beginning of the assembly task, and then characteristically dropped at approximately
\SI{0.25}{m}, which indicates that our model has learned the correspondence of this part of the geometry with the difficult initial
insertion phase. This hurdle is located where the edges of the active assembly cube collide with the side plates in the mockup.
The according plot of torques underlines this assumption: After correcting
the initial orientation, the torques peak again where the forces drop.
This shows the neural network's effort to get the
insertion right at this point, which it had generalized from the human
demonstrations.  Only when successfully passing this bottleneck was the
network more likely to apply forces of a higher magnitude, which nearly
vanish along with the torques upon reaching the goal position.  The results
show that our model has learned from human demonstrations to cope with jamming
situations for this specific task, and linking its predictions to the relative objects' geometry.

\subsection{Task to robot transfer}
\begin{figure*}[htbp] %
        \centering
        \includegraphics[width=0.98\textwidth]{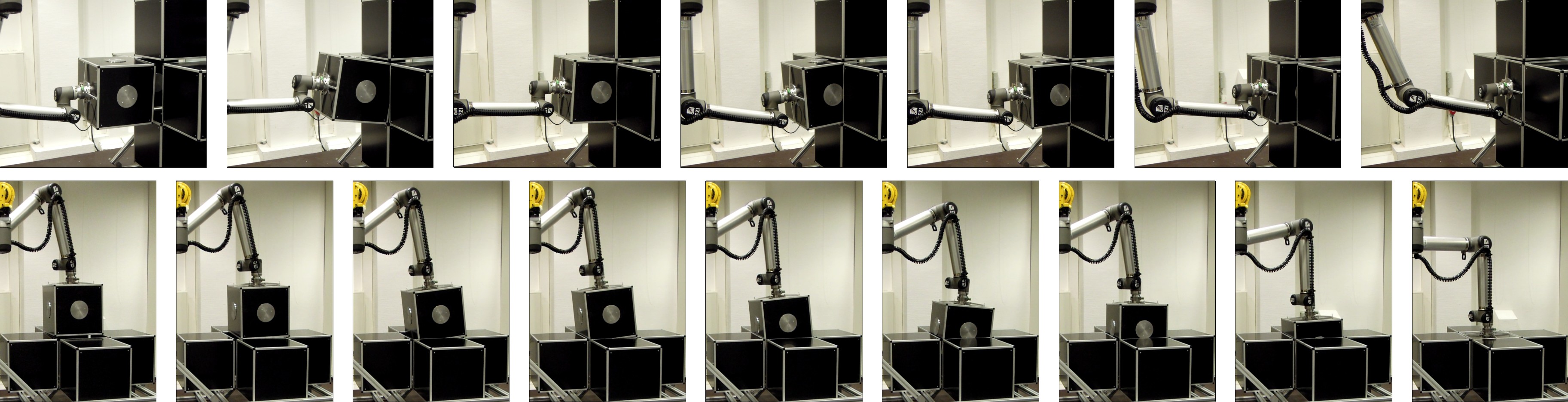}
        \caption{Skill execution on the UR10. Our experiments included 227 trials with randomly corrupted target poses.}
        \label{fig:skill_execution_ur10}
\end{figure*}

In this experiment, our goal was to transfer the learned skill to actual
hardware and evaluate its robustness with respect to an imprecise target.
To this end, we corrupted our ground truth ${}^b\bf{T}_t$ of the final assembly
with a random offset in the margin of \SI{50}{mm} linear
displacement and \SI{5}{deg} rotational displacement, which we applied along
random directions in a multitude of trials.
Figure~\ref{fig:skill_execution_ur10} shows an excerpt of the trials in different joint configurations for the robot.
During the runs we let our neural network predict
sequences of 50 steps, which we executed on the robot with our force control
during a constant time window of \SI{2.5}{s} for each sequence.
The force control was running at \SI{125}{Hz} (\SI{0.008}{s} cycle time, lower limit for the UR10), obtaining updates $\bm{f}^d$ to the force regulator every \SI{0.05}{s}.
For this experiment we constantly scaled the model's force output with a factor of $1.5$ and the torques with $2.0$ respectively.
We determined the magnitude in test runs prior to starting the evaluation.
Note that the ability to scale our model in this way provided an easy and
fast mechanism to fine-tune our skill to the real hardware.

\begin{figure}
        \centering
        \includegraphics[width=.5\textwidth]{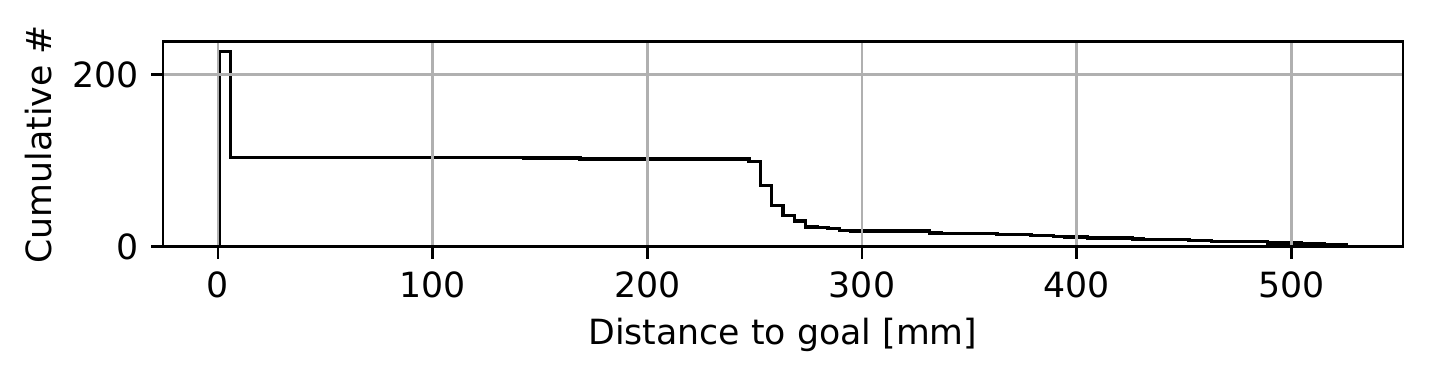}
        \caption{Cumulative histogram of trials, summing-up from right to left.}
        \label{fig:offset_performance_histogram}
\end{figure}

\begin{figure}
        \centering
        \includegraphics[width=.5\textwidth]{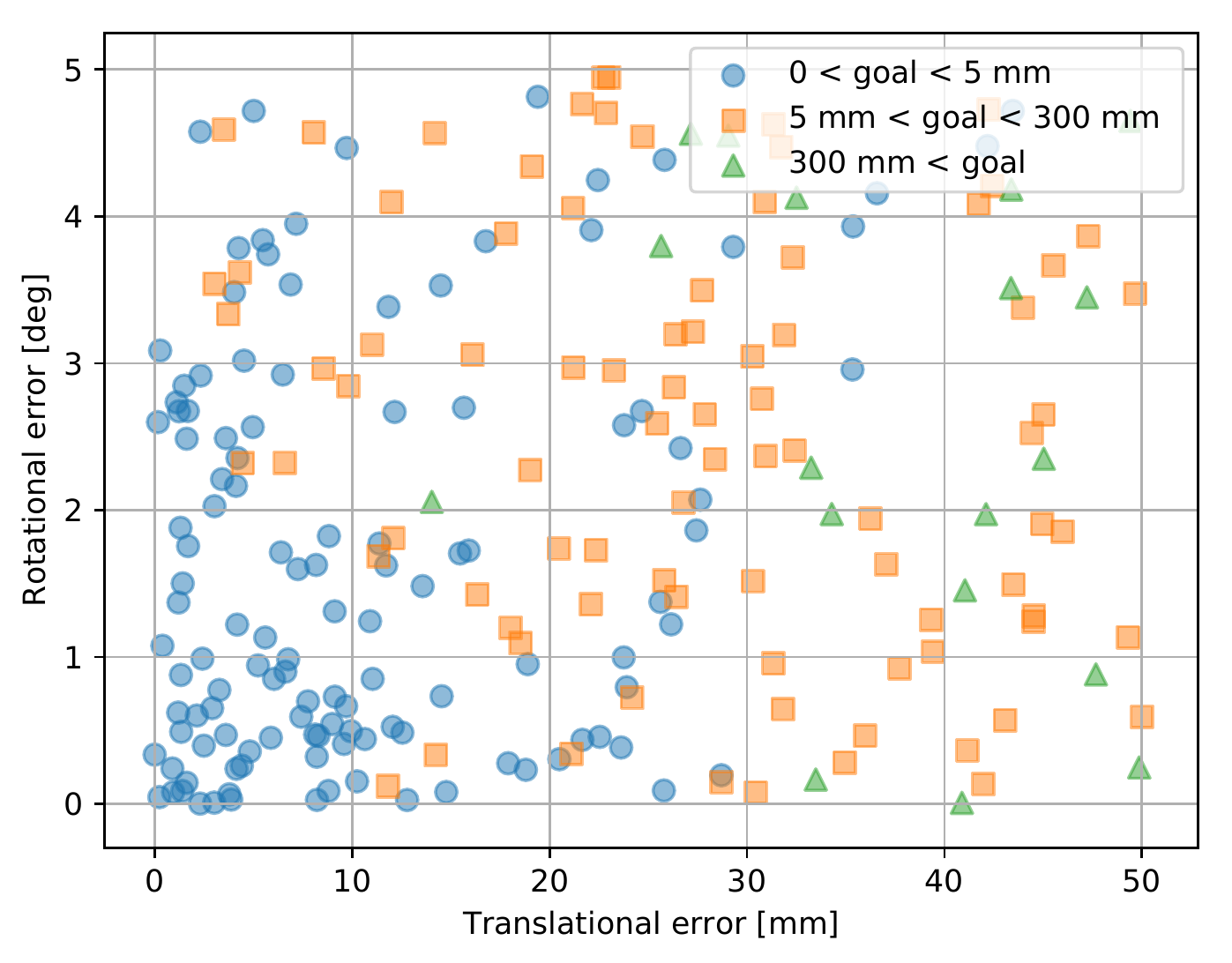}
        \caption{Performance of the skill with respect to localization errors of the setup.
        The target poses were corrupted randomly.
        The clearance between objects was \SI{2}{mm}
        and \SI{0.8}{grad}.
        The three classes categorize the individual executions,
        depending on distance to the ground truth goal pose. Blue circles
        represent successful trials.}
        \label{fig:offset_performance}
\end{figure}

Fig.~\ref{fig:offset_performance_histogram} shows the cumulative histogram (right to left) of the
trials vs. distance to the goal (ground truth).
We discontinued the individual executions when our skill model either made no further progress
or when we successfully reached the goal pose.
The curve shows that our model frequently got stuck at approximately
\SI{250}{mm} to the goal position, which relates to the decisive region, as was also experienced
with the forces and torques from Fig.~\ref{fig:jamming_forces} and Fig.~\ref{fig:jamming_torques}
respectively.
Note that all tries except for two were successful after surpassing this region, as
indicated by the almost horizontal line of Fig.~\ref{fig:offset_performance_histogram}.
Fig.~\ref{fig:offset_performance} shows the performance of our model for
random error combinations.
The blue circles represent successful executions.
As long as the errors do not exceed certain ranges, the results indicate a robust performance for up
to five times the linear clearance and three times the rotational clearance of the task.
For further increased errors, our model still succeeded in some cases, which
could be due to the random error direction being sometimes less severe.
The results show that our skill model for this task, although
purely trained on simulated data, was able to solve the task on the
real hardware, using a robot that was not part of the learning process.

Videos are included in the supplementary material.

\section{Discussion}
\label{sec:discussion}
\subsection{Robot control}
During the experiments on the real platform we anticipated that singularities of the
robot kinematics decreased the performance of the force controller.
This would also be the case for robots with limited morphology.
We propose to compute the kinematic manipulability in the workspace prior to executing the task and
position the robot accordingly.
Additionally, we consider articulated robots with a minimum of 6-axis for this purpose.
We also observed that the force-torque predictions of our model were partly compensated by
measurements of the force/torque sensor.
This was especially the case when the robot end-effector had rotated strongly from its taring position.
If sufficiently known, the masses and inertias of the active assembly object
could be gravity compensated during runtime for further improvements.

A control rate of \SI{125}{Hz} is commonly slow for stiff contacts.
To maintain contact stability with our setup, we
had to execute the task at low speeds.
Robots with faster read-write cycles might therefore be more suitable
for productive applications.

\subsection{Skills}
The time span of our network's memory covered only few seconds during training and inference.
Yet, we observed that the robot's execution on the setup followed several
longer-term strategies in the neighborhood of tens of seconds.
We address this effect to the chaining of small-scale, local sequences from the model's predictions that
together form strategies that are more complex over the course of execution.
By setting the friction conservatively high in simulation, we motivated the human experts
to avoid scratching along surfaces (because it slowed them down), and instead
to exploit clearance where possible.
Our results show that the demonstrations in simulation contained sufficient
contact-geometry semantics to solve our real-world task, such that our model
could overcome friction on the real setup without including force-torque
measurements in its input features.

\subsection{Scalability of the approach}
Although input feature scaling in our network provides a basic generalization for
the size of the objects, we assume limited
transferability to objects with highly different shapes.
This is due to the implicit encoding of geometry in the object-relative poses.
However, reusing the network's weights in combination with fine tuning for
another task is a promising direction for further research.

Finally, we want to mention that obtaining training data with the real robot in
teleoperation is also feasible with our approach, but implies more effort for
human experts in obtaining a similar number of labeled executions, albeit then
including real task parameters.
Note that although $1000$ samples seem much, this corresponds to approximately
three hours in simulation for obtaining a robot-transferable skill without requiring expertise in robot programming.

\section{CONCLUSIONS}   %
\label{sec:conclusions}
We presented a robot independent method to extract and learn
object-relative contact skills from human demonstrations in simulation,
using an LSTM-based neural network.
Our model learned to correct part tilting and jamming in contacts,
which we evaluated on a cube
assembly task on real hardware for different joint configurations.
Although solely trained on simulated data, the obtained skills were carried out robustly by
a Universal Robots UR10, which had not been included in the training process.
Excluding real force-torque sensor measurements from the network's input
features helped this transfer.
An advantage of our method is the ability to scale the neural
network's predictions without loosing their semantics:
Users can deploy their skills with reduced
magnitudes for safe test runs.

\section*{ACKNOWLEDGMENT}
We would like to thank Christoph Ganser for building the hardware demonstrator.
This work was supported in part by
the German Aerospace Center (DLR),
under national registration no. 50RA1503,
and the German Federal Ministry of Education and Research (BMBF) under funding code 16SV7715.

\renewcommand*{\bibfont}{\small}
\printbibliography

\end{document}